# Singularity Analysis of Lower-Mobility Parallel Manipulators Using Grassmann-Cayley Algebra

Daniel Kanaan, Philippe Wenger, Stéphane Caro and Damien Chablat

*Abstract*—This paper introduces a methodology to analyze geometrically the singularities of manipulators, of which legs apply both actuation forces and constraint moments to their moving platform. Lower-mobility parallel manipulators and parallel manipulators, of which some legs do not have any spherical joint, are such manipulators. The geometric conditions associated with the dependency of six Plücker vectors of finite lines or lines at infinity constituting the rows of the inverse Jacobian matrix are formulated using Grassmann-Cayley Algebra. Accordingly, the singularity conditions are obtained in vector form. This study is illustrated with the singularity analysis of four manipulators.

## I. INTRODUCTION

Parallel singularities are critical configurations in which a parallel manipulator loses its stiffness and gains some degree(s) of freedom (DOF). These singular configurations can be found using analytical, numerical or geometrical approaches [1-8]. Most research works are focused on 6-DOF manipulators in the literature, and particularly on Gough-Stewart parallel manipulators. For such manipulators, the rows of the inverse Jacobian matrix are Plücker coordinate vectors of lines. These lines are wrenches of actuation describing the instantaneous forces of actuation applied by the actuators to the moving platform. Their parallel singularities occur when lines within their legs are linearly dependent. Merlet analyzed the singularities of this type of manipulator using Grassmann line geometry [4]. Hao and McCarthy used screw theory to define conditions for line-based singularities [5]. They focused on 6-DOF parallel manipulators, of which each leg contains at least one actuator and the last three joints equivalent to a spherical joint. Consequently, each supporting leg of the system can apply only a pure force to the platform so that it is possible to characterize all singular configurations in terms of the geometry of linearly dependent sets of lines.

Ben-Horin and Shoham were one of the first to apply Grassmann-Cayley Algebra (GCA) to parallel manipulator singularities. In [6], [7], they analyzed parallel singularities of two classes of 6-DOF parallel manipulators using CGA. They used the superbracket decomposition and the Grassmann-Cayley operators to obtain the geometric conditions of singularities; namely, when four planes defined by the direction of the joint axes and the location of the spherical joint intersect in a common point. The manipulators analyzed in [6], [7] are not necessarily of the Gough-stewart type but their legs must transmit only pure forces.

Contrary to 6-DOF manipulators, the connectivity of the legs of a lower-mobility parallel manipulator is smaller than six and, in turn, each leg constrains partly the motion of the moving platform. When the legs of a manipulator lose their ability to constrain its moving platform, a so-called constraint singularity occurs. Joshi and Tsai developed a methodology to define the inverse Jacobian of lower-mobility parallel manipulators by means of the theory of reciprocal screws [8], i.e., a 6x6-inverse Jacobian matrix is defined, of which rows are wrenches providing information about both architecture and constraint singularities. Those wrenches, also known as governing lines, are the actuation forces and constraint moments applied to the moving platform.

In this paper, we focus on the identification of parallel singularities of lower-mobility parallel manipulators by means of GCA. Thus, the GCA is applied to parallel manipulators, of which the rows of their 6x6 inverse Jacobian matrix can be finite lines (zero pitch screws) or lines at infinity (infinite pitch screws).

It turns out that previous studies focused only on the singularity analysis of parallel manipulators, of which legs apply pure forces on the moving platform. It means that the parallel singularities of such manipulators are related to the dependency between finite lines. Consequently, the previous geometrical methods can be only applied to manipulators, of which legs have at least one passive spherical joint. Moreover, GCA based methods previously proposed did not analyze manipulators with spatial parallelograms as these methods do not handle lines that remain always parallel, i.e. intersect at infinity. The methodology introduced in this paper allows us to identify the parallel singularities related to the dependency between finite lines and lines at infinity in the projective space. Accordingly, the geometric methods can be applied to a wider range of parallel manipulators as the manipulators, of which legs apply both actuation forces and constraint

This work was supported partly by the French Research Agency A.N.R. (Agence Nationale pour la Recherche) and by the Integrated Project NEXT of the European Community. The authors are with the Institut de Recherche en Communications et Cybernétique de Nantes (IRCCyN), 1, rue de la Noë 44321 Nantes, France. Email: Philippe.Wenger@ irccyn.ec-nantes.fr

moments to their moving platform. It includes lower mobility parallel manipulators, parallel manipulators, of which some legs do not have any spherical joint, and parallel manipulators with spatial parallelograms.

This paper is organized as follows: Section II presents briefly the bracket ring, the GCA operators and the Superbracket decomposition, which is equivalent to the determinant of the inverse Jacobian matrix. Section III introduces the projective space and its geometric elements that can be either finite or infinite. It also sums up the basic concepts of screws that carry the wrenches acting to the moving platform, the latter being actuation or constraint forces, as known as zero pitch screw, and constraint moments as known as infinite pitch screw, in the 3-dimensional projective space. Section IV introduces the singularity geometric conditions of three classes of lower-mobility parallel manipulators. Finally, the singularities of four manipulators are analyzed in Section V as illustrative examples, namely, *i*) the 3-UPU parallel manipulator; *ii*) the Delta-linear robot; *iii*) the McGill Schönflies Motion Generator (McGill SMG); and *iv*) the parallel module of the Verne machine.

## II. GRASSMANN-CAYLEY ALGEBRA

Within the framework of this study, the GCA is used to transform the singular geometric conditions defined as the dependency between six lines expressed in the 3-dimensional projective space $\mathsf{P}^3$, in coordinate free algebraic expressions involving twelve points selected on the axes of these lines. Only the basic concepts of GCA are recalled. For further details on GCA, the reader is referred to [7], [9-14] and references therein.

### A. Basic elements and operations

The basic elements of GCA are called *extensors*, which in fact are symbolically denoted by Plücker coordinates of vectors. The extensors are vectors that represent geometric entities and are characterized by their *step*. Extensors of step 1, 2 and 3 stand for a point, a line and a plane, respectively. Special determinants called *brackets* are also defined in GCA. Let us consider a finite set of points $\{e_1, e_2,..,e_d\}$ defined in the *(d-1)*-dimensional vector space, $V$, over the field $\Box$. Each point $e_i$ is represented by a *d*-tuple using homogeneous coordinates with $e_i = e_{1,i}, e_{2,i},..., e_{d,i}$ ($1 \leq i \leq d$). The bracket of these points is defined as the determinant of the matrix, of which columns are the homogeneous coordinates of points $e_i$ ($1 \leq i \leq d$):

$$[e_1, e_2,...,e_d] = \begin{vmatrix} e_{1,1} & e_{1,2} & ... & e_{1,d} \\ \vdots & \vdots & ... & \vdots \\ e_{d,1} & e_{d,2} & ... & e_{d,d} \end{vmatrix} \quad (1)$$

If $e_i = e_j$ with $i \neq j$, or $e_1, e_2, ..., e_d$ are dependent

$$[e_1, e_2,...,e_d] = 0 \quad (2)$$

For any permutation $\sigma$ of $\{1,2,...,d\}$

$$[e_1, e_2,...,e_d] = sign(\sigma)[e_{\sigma 1}, e_{\sigma 2},..., e_{\sigma d}] \quad (3)$$

$$[e_1, e_2,...,e_d][f_1, f_2,...,f_d] = \sum_{j=1}^{d}[f_j, e_2,...,e_d][f_1, f_2,...,f_{j-1}, e_1, f_{j+1},...,f_d] \quad (4)$$

Two basic operations involving extensors play an essential role in GCA: the *join* and *meet* operators. The *join* is associated with the union of two vector spaces while the *meet* has the same geometric meaning as the intersection of two vector spaces. Let $\mathbf{A} = a_1 \vee a_2 \vee ... \vee a_i = a_1 a_2 ... a_i$ and $\mathbf{B} = b_1 \vee ... \vee b_j = b_1 b_2 ... b_j$ be two extensors of step *i* and *j*, respectively, defined in the *(d-1)*-projective space, their join $\mathbf{A} \vee \mathbf{B}$ is an extensor of step *i+j* defined as follows:

$$\mathbf{A} \vee \mathbf{B} = a_1 \vee a_2 \vee ... \vee a_i \vee b_1 \vee b_2 \vee ... \vee b_j = a_1 a_2 ... a_i b_1 b_2 ... b_j \quad (5)$$

It is not null as long as vectors $\{a_1, a_2,..., a_i, b_1, b_2,..., b_j\}$ are linearly independent. If $i+j \geq d$, the meet $\mathbf{A} \wedge \mathbf{B}$ is an extensor of step *i+j-d* defined as:

$$\mathbf{A} \wedge \mathbf{B} = \sum_{\sigma}\left[a_{\sigma(1)} a_{\sigma(2)} ... a_{\sigma(d-j)} b_1 b_2 ... b_j\right] a_{\sigma(d-j+1)}...a_{\sigma(i)} \quad (6)$$

where the sum is taken over all permutations $\sigma$ of $\{1,2,...,i\}$ such that $\sigma(1) < \sigma(2) < ... < \sigma(d-j)$ and $\sigma(d-j+1) < \sigma(d-j+2) < ... < \sigma(i)$. A shorter notation [10], which will be used here, is to write a dot above the permuted elements instead of using $\sigma$ with the summation and $sign(\sigma)$ implicit:

$$\mathbf{A} \wedge \mathbf{B} = \left[\dot{a}_1 \dot{a}_2 ... \dot{a}_{d-j} b_1 b_2 ... b_j\right] \dot{a}_{d-j+1}...\dot{a}_i \quad (7)$$

### B. Examples

We now give three examples of how to obtain simplified expressions of the meet operation between entities. These examples will be useful in section IV. The first two ones (meet of two and four 3-extensors, respectively) were already obtained in [7]. The third one was recently derived in [15].

The meet $\mathbf{A} \wedge \mathbf{B}$ of two extensors of step 3 ($i=j=3$) representing two non-coplanar planes in the projective space $\mathsf{P}^3$ ($d=4$) is an extensor of step $i+j-d = 2$ representing the intersection line of the two planes (possibly at infinity). It can be written as [7]:

$$\mathbf{A} \wedge \mathbf{B} = (abc)(def) = [adef]bc - [bdef]ac - [cdef]ba = \left[\dot{a}\,def\right]\dot{b}\dot{c} \quad (8)$$

Let $\mathbf{D}=ijk$ and $\mathbf{E}=lmn$ represent two more non-coplanar planes in $\mathsf{P}^3$. The meet $(\mathbf{A} \wedge \mathbf{B}) \wedge (\mathbf{D} \wedge \mathbf{E})$ is an extensor of step 0 defined by [7]:

$$(\mathbf{A} \wedge \mathbf{B}) \wedge (\mathbf{D} \wedge \mathbf{E}) = \left[\dot{a}\,def\right]\left[\dot{b}\,ijk\right]\left[\dot{c}\,lmn\right] \quad (9)$$

It represents the intersection of four planes in the projective space $\mathsf{P}^3$.

We now want to calculate a simplified expression of the meet $(\mathbf{A} \wedge \mathbf{B}) \wedge \mathbf{C}$, where $\mathbf{C}$ is an extensor of step 2 representing a line. It will be used to interpret geometrically the singularities of lower-mobility manipulators with 6 governing lines (section IV-D). By definition, the meet

$(\mathbf{A} \wedge \mathbf{B}) \wedge \mathbf{C}$ is an extensor of step 0 that takes the following form:

$$(\mathbf{A} \wedge \mathbf{B}) \wedge \mathbf{C} = \left[\dot{a}\,def\right]\left[\ddot{b}\,\ddot{c}\,gh\right] = [adef][bcgh] - [bdef][acgh] - [cdef][bagh] \quad (10)$$

Using the bracket property (4), it turns out that:

$$[adef][bcgh] = [bdef][acgh] + [cdef][bagh] + [gdef][bcah] + [hdef][bcga] \quad (11)$$

Accordingly, we obtain the following expression by combining (10) and (11):

$$(\mathbf{A} \wedge \mathbf{B}) \wedge \mathbf{C} = [gdef][bcah] + [hdef][bcga]$$
$$= [defg][habc] - [defh][gabc] = \left[def\,\dot{g}\right]\left[\dot{h}\,abc\right] \quad (12)$$

*C. Superbracket Decomposition*

The rows of the inverse Jacobian matrix of a parallel manipulator are the Plücker coordinates of six lines in the projective space $\mathsf{P}^3$. The superjoin of the six corresponding vectors in $\mathsf{P}^5$ corresponds to the determinant of their six Plücker coordinate vectors up to a scalar multiple, which is the *superbracket* in GCA $\Lambda(V^{(2)})$ [13]. Thus, a singularity occurs when these six Plücker coordinate vectors are dependent, which is equivalent to a superbracket equal to zero. White [13] and McMillan [14] used the theory of projective invariants to decompose the superbracket into an expression having brackets involving twelve points selected on those lines. This expression can be transformed into a linear combination of 24 bracket monomials, each bracket monomial being the product of three brackets:

$$[ab, cd, ef, gh, ij, kl] = \sum_{i=1}^{24} y_i \quad (13)$$

The reader is reported to [7,14] and Appendix A for the expression of these 24 monomials.

### III. PROJECTIVE SPACE AND SCREWS

*A. Infinite elements in the 3-dimensional projective space*

In projective geometry, the plane at infinity, $\Pi_\infty$, is a projective plane that is added to the affine space, $\square^3$, in order to give it closure of incidence properties. The result of the addition is the three-dimensional projective space, $\mathsf{P}^3$. Any pair of parallel lines in $\mathsf{P}^3$ intersects each other at a point in $\Pi_\infty$, which is called a point at infinity. Also, every line in $\mathsf{P}^3$ intersects $\Pi_\infty$ at a unique point at infinity. This point is only determined by the direction of the line. Any pair of parallel planes in $\mathsf{P}^3$ will intersect each other at a projective line in $\Pi_\infty$, which is called a line at infinity. Also, every plane in $\mathsf{P}^3$ intersects $\Pi_\infty$ at a unique line at infinity. This line is only determined by the vector normal to the corresponding plane.

As a matter of fact, $\Pi_\infty$ consists of all points at infinity of the projective space and contains all lines at infinity. This plane is obtained by the definition of three points at infinity that are not aligned. Therefore, a line at infinity passes through two points at infinity and consists of all points at infinity belonging to the same family of parallel planes.

Furthermore, any two non-parallel planes intersect each other at a line in $\mathsf{P}^3$. Besides, this line intersects $\Pi_\infty$ at a point at infinity. This point belongs to the two non-parallel planes as well as to their corresponding lines at infinity. Consequently any two lines at infinity intersect each other in $\Pi_\infty$.

In the scope of this study, the concept of infinite elements in $\mathsf{P}^3$ allows us to represent forces and moments with lines and to deal with their corresponding relations. This is the key element of the new contribution of this paper: points and lines at infinity make it possible to represent constraint moments and pure forces with parallel lines.

*B. Reciprocal Screws applications*

Screws are useful to represent the motion of joints constituting the serial limbs as well as their reciprocal screws, which are wrenches applied by these serial limbs to the moving platform. These wrenches, stated as governing lines of a parallel manipulator, constitute the rows of the inverse Jacobian matrix. In this subsection, the reciprocal screw theory is summed up to help the readers better understand screws and their applications to obtain the governing lines of a parallel manipulator. Two screws, representing the twist, **T**, and the wrench, **W**, respectively, are reciprocal, if the wrench acts on a rigid body without producing any work while the body undergoes an infinitesimal twist. It means that these screws are reciprocal as long as the virtual work of the wrench **W** acting on the twist **T** is null.

In case both screws are lines or lines at infinity in the 3-dimensional projective space, they become reciprocal when they intersect, are collinear or parallel.

The line of a zero pitch screw reciprocal to the twist screw of a prismatic joint is perpendicular to the sliding direction. In other words, the corresponding lines of both screws intersect.

All screws that are reciprocal to a *n*-system screw ($n < 6$) form a $(6-n)$-system, [17]. Furthermore a *n*-screw system can be replaced by another equivalent *n*-screw system by applying a linear transformation to the basis of the first one [19]. For more details on reciprocal screw theory, the reader is referred to [5], [8], [16-18].

### IV. SINGULARITY GEOMETRIC CONDITIONS OF LOWER-MOBILITY PARALLEL MANIPULATORS

*A. Three classes of parallel manipulators*

This study aims at broadening the application of line geometry theory in order to analyze the singularities of lower-mobility parallel manipulators. It is noteworthy that the legs of these manipulators can apply both actuation and constraint wrenches to the moving platform, and thus, they do not necessarily have a passive spherical joint. Consequently, the rows of their 6x6-inverse Jacobian matrix are Plücker coordinate vectors of a finite line or a line at infinity. The singular configurations occur when the governing lines constituting the inverse Jacobian matrix are linearly dependent, which is equivalent to a superbracket equal to zero. In this section, we use GCA operators and the superbracket decomposition to determine the geometric singularity conditions of three classes of parallel manipulators defined according to the features of their governing lines. These governing lines are associated with the kind of wrenches applied to the moving platform, namely,

- class I: Three actuation forces and three constraint moments (Fig. 1);
- class II: Two pairs of intersecting actuation forces and two constraint moments (Fig. 2);
- class III: Six actuation forces such that at least two pairs of governing lines are parallel (Fig. 3).

### B. Manipulators with three actuation forces and three constraint moments

Manipulators belonging to the first class are characterized by a moving platform subject to three actuation forces (zero pitch screw), i.e., $\hat{\mathbf{F}}_i = \begin{bmatrix} \mathbf{s}_i^T & (\mathbf{r}_i \times \mathbf{s}_i)^T \end{bmatrix}^T$ ($i$=1,2,3), and three constraint moments (infinite pitch screw), i.e., $\hat{\mathbf{M}}_i = \begin{bmatrix} 0_{1\times 3} & \mathbf{n}_i^T \end{bmatrix}^T$ ($i$=1,2,3), $\mathbf{s}_i$ being a unit vector in the direction of the line of application of actuation force $\hat{\mathbf{F}}_i$, $\mathbf{r}_i$ being the position vector of a point on this line and $\mathbf{n}_i$ the direction of the torque associated with constraint moment $\hat{\mathbf{M}}_i$. The rows of the 6x6 inverse Jacobian matrix are the Plücker coordinate vectors of three lines (actuation forces) and three lines at infinity (constraint moments). The dependency between these lines is related to the degeneration of the inverse Jacobian matrix, which is equivalent to a superbracket equal to zero.

Let $a\underline{b}$, $c\underline{d}$, $e\underline{f}$ be the lines representing the three actuation forces $\mathbf{F}_i$ ($i$=1,2,3), $a$, $c$, $e$ being points and $\underline{b}$, $\underline{d}$, $\underline{f}$ points at infinity, the underlined letter characterizing a point at infinity. Points $\underline{b}$, $\underline{d}$, $\underline{f}$ are chosen at infinity and not on the moving platform as, by doing so, simplifications in the bracket expression are more likely to arise in (13) because points at infinity are always coplanar. These points are expressed by means of their homogeneous coordinates, i.e., $\underline{b} = (\mathbf{s}_1^T \ \mathbf{0})^T$, $\underline{d} = (\mathbf{s}_2^T \ \mathbf{0})^T$, $\underline{f} = (\mathbf{s}_3^T \ \mathbf{0})^T$. On the other hand, since every line at infinity meets every other line at infinity, the three constraint moments $\mathbf{M}_i$ ($i$=1, 2, 3) can be represented by the three lines at infinity $\underline{gh}$, $\underline{gi}$ and $\underline{hi}$, respectively where points $\underline{g}$, $\underline{h}$ and $\underline{i}$ are defined by their vectors $\mathbf{g} = \mathbf{n}_1 \times \mathbf{n}_2$, $\mathbf{h} = \mathbf{n}_1 \times \mathbf{n}_3$ and $\mathbf{i} = \mathbf{n}_2 \times \mathbf{n}_3$, respectively.

From (13) and due to the repetition of points in the same bracket, we can simplify the superbracket expression into a reduced number of non-zero monomial terms. Therefore, the superbracket decomposition of this type of manipulators becomes:

$$\begin{bmatrix} a\underline{b}, \ c\underline{d}, \ e\underline{f}, \ \underline{gh}, \ \underline{gi}, \ \underline{hi} \end{bmatrix} = \begin{bmatrix} a\underline{bdf} \end{bmatrix}\begin{bmatrix} c\underline{ghi} \end{bmatrix}\begin{bmatrix} e\underline{ghi} \end{bmatrix} \qquad (14)$$

with

$$\begin{bmatrix} a\underline{bdf} \end{bmatrix} = (\mathbf{s}_1 \times \mathbf{s}_2) \cdot \mathbf{s}_3 \qquad (15a)$$

and

$$\begin{bmatrix} e\underline{ghi} \end{bmatrix} = \begin{bmatrix} c\underline{ghi} \end{bmatrix} = \begin{bmatrix} (\mathbf{n}_1 \times \mathbf{n}_2) \cdot \mathbf{n}_3 \end{bmatrix}^2 \qquad (15b)$$

Consequently, the manipulator under study meets a singularity whenever $(\mathbf{s}_1 \times \mathbf{s}_2) \cdot \mathbf{s}_3 = 0$ or $(\mathbf{n}_1 \times \mathbf{n}_2) \cdot \mathbf{n}_3 = 0$, i.e. when the area of triangles $\underline{ghi}$ or $\underline{bdf}$ is null. These two conditions include cases (b-g) in Fig. 1. Each vertex of triangle $\underline{bdf}$ is a point at infinity representing the direction of an actuation force. In cases (b,c), the three points at infinity $\underline{b}$, $\underline{d}$ and $\underline{f}$ are collinear, so three vectors $\mathbf{s}_i$ are coplanar. Cases (c,d) occur when at least two points at infinity coincide, i.e., when two vectors $\mathbf{s}_i$ are parallel. Each side of triangle $\underline{ghi}$ is a line at infinity defined by the intersection of a family of parallel planes at infinity, $\mathbf{n}_i$ being the normal to these planes. Cases (e,f) occur when the three lines at infinity intersect, i.e, when vectors $\mathbf{n}_i$ are coplanar. In cases (f,g), at least two lines at infinity are collinear, which means that their corresponding planes are parallel so at least two vectors $\mathbf{n}_i$ are parallel. Let us notice that another condition may appear when at least one constraint moment degenerates, meaning that at least one line $\underline{gh}$, $\underline{gi}$ or $\underline{hi}$ degenerates to a point. This condition is more related to the arrangement of the joint within each leg.

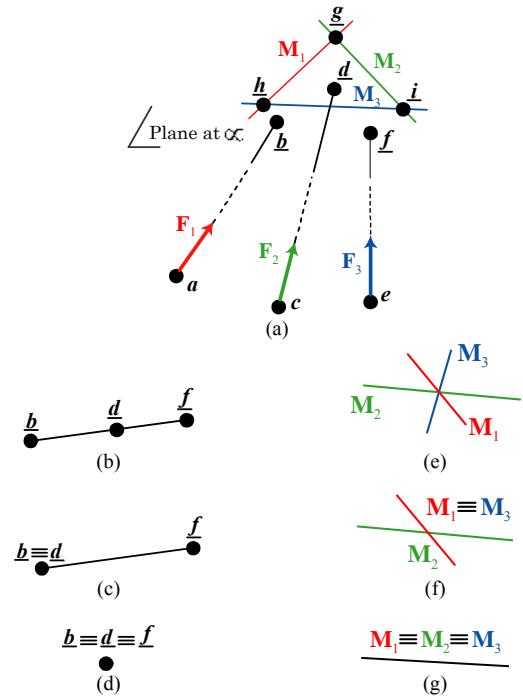

Figure 1: Geometric conditions for parallel singularities of manipulators with three actuation forces and three constraint moments

## C. *Manipulators with two pairs of intersecting actuation forces and two constraint moments*

Manipulators belonging to the second class are characterized by a moving platform subject to two pairs of intersecting forces (zero pitch screw) and two constraint moments (infinite pitch screw).

Let *ab*, *cb*, *de*, *fe* be the lines representing the four forces $\hat{\mathbf{F}}_{ij}$ ($i=1,2$, $j=1,2$) and let $\underline{gh}$, $\underline{ih}$, be the lines at infinity representing the two constraint moments $\hat{\mathbf{M}}_i = \begin{bmatrix} 0_{1\times 3} & \mathbf{n}_i^T \end{bmatrix}^T$ ($i=1,2$), respectively, $\mathbf{n}_i$ being the direction of the torque associated with constraint moment $\hat{\mathbf{M}}_i$. From (13) and due to the repetition of points in the same bracket, the superbracket expression becomes:

$$[ab, cb, de, fe, \underline{gh}, \underline{ih}] =$$
$$[abcd][bfe\underline{h}][eg\underline{ih}] - [abce][bfe\underline{h}][dg\underline{ih}] - \quad (16)$$
$$[abcf][bde\underline{h}][eg\underline{ih}] + [abce][bde\underline{h}][fg\underline{ih}]$$

After collecting equal brackets, applying the shuffles relation of (6) and permuting *d*, *e* and *f*, we obtain:

$$[ab, cb, de, fe, \underline{gh}, \underline{ih}] = [abc\overset{\square}{d}][b\overset{\square}{f}e\underline{h}][\overset{\square}{e}\underline{gih}] \quad (17)$$

The right hand side of (17) is the result of the meet of the four planes, (*abc*), (*dfe*), (*beh*) and (*gih*) as shown in (7). The last plane being at infinity, its intersections with the other three planes are three lines at infinity, $L_i = \begin{bmatrix} 0_{1\times 3} & \mathbf{m}_i^T \end{bmatrix}^T$ ($i=1,2,3$), respectively, $\mathbf{m}_i$ being the vector normal to the corresponding plane. The singular geometric condition of this class of parallel manipulators includes the following cases: (*i*) one of the planes degenerates into a line, $\|\mathbf{n}_1 \times \mathbf{n}_2\|\|\mathbf{m}_1\|\|\mathbf{m}_2\|\|\mathbf{m}_3\| = 0$ (*ii*) the four planes intersect in a common point. This case appears when triangle *jkl* vanishes, where *j*, *k* and *l* are defined by their vectors $\mathbf{m}_1 \wedge \mathbf{m}_3$, $\mathbf{m}_1 \wedge \mathbf{m}_2$ and $\mathbf{m}_2 \wedge \mathbf{m}_3$, respectively, as shown in Fig. 2. Finally, it turns out that the singular condition is written as follows:

$$(\mathbf{m}_1 \times \mathbf{m}_2) \cdot \mathbf{m}_3 = 0 \quad (18)$$

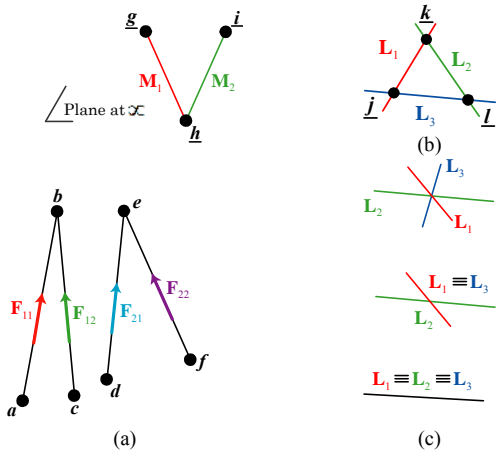

Figure 2: Geometric conditions for parallel singularities of manipulators with two pairs of intersecting actuation forces and two constraint moments

## D. *Manipulators with six actuation forces such that at least two pairs of governing lines are parallel*

Manipulators belonging to the third class are characterized by a moving platform subject to 6 forces (zero pitch screw) with at least two pairs having parallel supporting lines.

Let *ab*, *cd*, *ef*, *gh*, *ij kl*, be the lines representing the six forces $\mathbf{F}_{ij}$ ($i=1,2,3$, $j=1,2$), respectively, with points *a*, *c*, *e*, *g*, *i*, *k* being the intersections of the lines and the base, and *b*, *d*, *f*, *h*, *j*, *l* being the intersection of the lines and the platform. We suppose that lines *ef* and *gh* are parallel and intersect at infinity. Likewise, lines *ij*, *kl* are parallel and intersect at infinity. The superbracket decomposition of these lines is then reduced to:

$$[ab, cd, ef, gh, ij, kl] = [a\underline{m}, c\underline{n}, e\underline{o}, g\underline{o}, i\underline{p}, k\underline{p}] \quad (19)$$

where $\underline{m}$, $\underline{n}$, $\underline{o}$ and $\underline{p}$ are defined by their vectors $\mathbf{m} = \mathbf{b} \square \mathbf{a}$, $\mathbf{n} = \mathbf{d} \square \mathbf{c}$, $\mathbf{o} = \mathbf{f} \square \mathbf{e} = \mathbf{h} \square \mathbf{g}$ and $\mathbf{p} = \mathbf{j} \square \mathbf{i} = \mathbf{l} \square \mathbf{k}$.

The shortest form of the superbracket decomposition using the algorithm defined in [23] results in the following non-zero monomial terms:

$$[\underline{oegm}][\underline{oncp}][ai\underline{p}k] - [\underline{oega}][\underline{oncp}][mi\underline{p}k] - \quad (20)$$
$$[\underline{oegn}][\underline{omap}][ci\underline{p}k] + [\underline{oegc}][\underline{omap}][ni\underline{p}k] = 0$$

After collecting equal brackets, applying the shuffles relation of (6), and permuting $\underline{m}$, *a* and $\underline{n}$, *c*, we obtain:

$$[\underline{oncp}][\underline{oeg}\overset{\square}{m}][\overset{\square}{a}i\underline{p}k] - [\underline{omap}][\underline{oeg}\overset{\square}{n}][\overset{\square}{c}i\underline{p}k] = 0 \quad (21)$$

Applying the corresponding relation:

$$[(b-a)cde] = [bcde] - [acde] \quad (22)$$

we can replace some points at infinity by points and obtain the following expression:

$$[ab, cd, ef, gh, ij, kl] =$$
$$[\underline{oncp}][\overset{\square}{feg}\,b][\overset{\square}{a}\,ijk] - [\underline{omap}][\overset{\square}{feg}\,d][\overset{\square}{c}\,ijk] \quad (23)$$

By using (12), we can prove that

$$[\overset{\square}{feg}\,b][\overset{\square}{a}\,ijk] = (ijk \wedge feg) \wedge ba = tu \wedge ba = [tuba] \quad (24a)$$

and

$$[\overset{\square}{feg}\,d][\overset{\square}{c}\,ijk] = (ijk \wedge feg) \wedge dc = tu \wedge dc = [tudc] \quad (24b)$$

*tu* being the intersecting line of planes including legs II and III, i.e., $\mathbf{tu} = (\mathbf{ef} \times \mathbf{eg}) \times (\mathbf{ij} \times \mathbf{ik})$

On the other hand, it is easy to prove that:

$$[\underline{oncp}] = [fe\underline{pn}] = [feqr] \text{ and } [\underline{omap}] = [feqs] \quad (25)$$

where $\underline{q}$, $\underline{r}$ and $\underline{s}$ are defined by their vectors $\mathbf{q} = \mathbf{e} + \mathbf{p}$, $\mathbf{r} = \mathbf{q} + \mathbf{n}$ and $\mathbf{s} = \mathbf{q} + \mathbf{m}$.

Thus, the invariant algebraic expression related to the existence of parallel singularities of the manipulator under study can be stated as:

$$[feqr][tuba] - [feqs][tudc] = 0 \quad (26)$$

where $[tuba] = \mathbf{tu} \cdot (\mathbf{ub} \times \mathbf{ab})$ and $[tudc] = \mathbf{tu} \cdot (\mathbf{ud} \times \mathbf{cd})$ with $\mathbf{tu} = (\mathbf{ef} \times \mathbf{eg}) \times (\mathbf{ij} \times \mathbf{ik})$ whereas $[feqr] = \mathbf{cd} \cdot \mathbf{N}$ with $\mathbf{N} = \mathbf{ef} \times \mathbf{ij}$ and $[feqs] = \mathbf{ab} \cdot \mathbf{N}$.

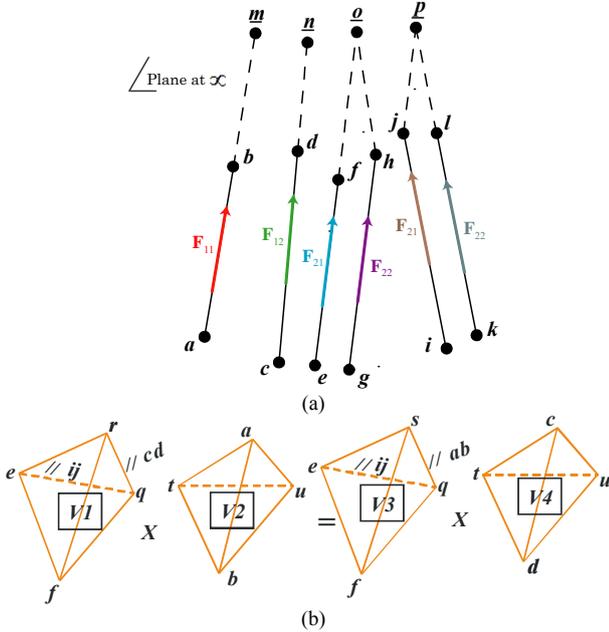

Figure 3: Geometric conditions for parallel singularities of manipulators with six actuation forces such that at least two pairs of governing lines are parallel

The previous expression is geometrically equivalent to the difference between the volume products of two pairs of tetrahedrons with vertices expressed as a function of points *a,b,...l* as shown in Fig. 3b. This geometric condition includes the following cases: (*i*) planes of two lines having spatial parallelogram shape are coplanar or parallel so $\|\mathbf{tu}\| = 0$; (*ii*) *ef* and *ij* are parallel so $\|\mathbf{N}\| = 0$; (*iii*) *ef*, *cd* are parallel and *ij*, *ab* are parallel or *ef*, *ab* are parallel and *ij*, *cd* are parallel; (*iv*) *ab* and *cd* intersect with *tu*, in this case the six actuation forces form a singular linear complex; (*v*) *ef*, *cd* are parallel and *tu*, *cd* are coplanar or *ij*, *cd* are parallel and *tu*, *cd* are coplanar or *ij*, *ab* are parallel and *tu*, *ab* are coplanar or *ef*, *ab* are parallel and *tu*, *ab* are coplanar; (*vi*) the six actuation forces form a general linear complex and the singularity conditions is:

$$[\mathbf{cd} \cdot \mathbf{N}][\mathbf{tu} \cdot (\mathbf{ub} \times \mathbf{ab})] - [\mathbf{ab} \cdot \mathbf{N}][\mathbf{tu} \cdot (\mathbf{ud} \times \mathbf{cd})] = 0 \quad (27)$$

It is interesting to note that this condition is equivalent to the one recently introduced in [15] for robots of class *c1c1a2a2*, which have finite lines of action.

## V. ILLUSTRATIVE EXAMPLES

In order to illustrate the results obtained in Section IV, we analyze the singularities of four manipulators:

1. The 3-UPU manipulator;
2. the Delta-linear manipulator;
3. the McGill SMG;
4. the parallel module of the Verne machine.

### A. Singularity analysis of the 3-UPU manipulator

The 3-UPU manipulator, shown in Fig. 4, was studied in [3], [8], [21], [24].

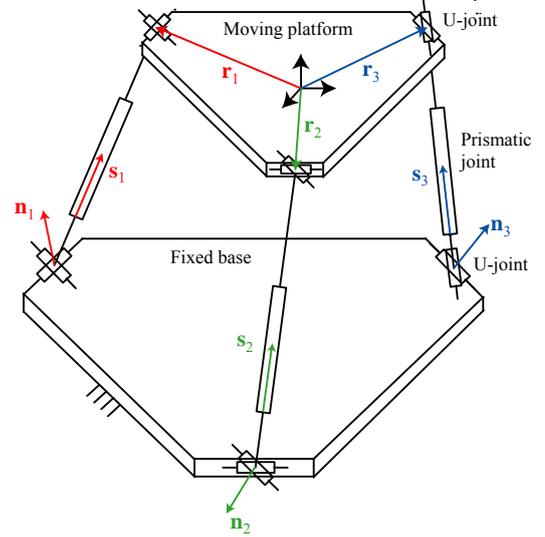

Figure 4: The 3-UPU manipulator

The moving platform is controlled by means of three linear actuators mounted on three identical legs and the translational motion is obtained with a special arrangement of the U joints[1]. Each leg has a connectivity equal to five. Accordingly, it applies one actuation force $\hat{\mathbf{F}}_i = \begin{bmatrix} \mathbf{s}_i^T & (\mathbf{r}_i \times \mathbf{s}_i)^T \end{bmatrix}^T$ ($i$=1,2,3), and one constraint moment $\hat{\mathbf{M}}_i = \begin{bmatrix} \mathbf{0}_{1\times 3} & \mathbf{n}_i^T \end{bmatrix}^T$ to the moving platform. $\mathbf{s}_i$ is a unit vector in the direction of leg $i$. $\mathbf{r}_i$ is the position vector of a point on this leg. $\mathbf{n}_i$ is the cross product of the two vectors associated with the first U joint axes of leg $i$. It represents also the direction of the torque associated with the constraint moment $\hat{\mathbf{M}}_i$. Each actuation force $\hat{\mathbf{F}}_i$ is a zero pitch screw reciprocal to all joint screws[2] of leg $i$ except to the joint screw associated with the actuated prismatic joint of the same leg. Each constraint moment $\hat{\mathbf{M}}_i$ is an infinite pitch screw reciprocal to all joint screws of leg $i$. Consequently, the transpose of the inverse Jacobian of the 3-UPU can be expressed as follows [8],

$$\mathbf{J}^{-T} = \begin{bmatrix} \mathbf{s}_1 & \mathbf{s}_2 & \mathbf{s}_3 & \mathbf{0}_{1\times 3} & \mathbf{0}_{1\times 3} & \mathbf{0}_{1\times 3} \\ \mathbf{r}_1 \times \mathbf{s}_1 & \mathbf{r}_2 \times \mathbf{s}_2 & \mathbf{r}_3 \times \mathbf{s}_3 & \mathbf{n}_1 & \mathbf{n}_2 & \mathbf{n}_3 \end{bmatrix} \quad (28)$$
$$= \begin{bmatrix} \hat{\mathbf{F}}_1 & \hat{\mathbf{F}}_2 & \hat{\mathbf{F}}_3 & \hat{\mathbf{M}}_1 & \hat{\mathbf{M}}_2 & \hat{\mathbf{M}}_3 \end{bmatrix}$$

Due to the form of its inverse Jacobian matrix, the 3-UPU manipulator can be considered topologically equivalent to a manipulator of class I. Thus, it belongs to the first class of parallel manipulators.

From Section IV.B, *ab*, *cd*, *ef*, are the lines representing the three forces $\hat{\mathbf{F}}_i$ ($i$=1,2,3) and *gh*, *gi* and *hi*, are lines at infinity representing the three constraint moments $\mathbf{M}_i$ ($i$=1,2,3). Therefore, the 3-UPU manipulator meets a singularity whenever ($\mathbf{s}_1 \times \mathbf{s}_2$).$\mathbf{s}_3$=0, which occurs in Figs. 1 (b-d). On the other hand, the manipulator reaches a constraint singularity whenever ($\mathbf{n}_1 \times \mathbf{n}_2$).$\mathbf{n}_3$=0, which occurs in Figs. 1 (e-g). Let us notice that another condition may appear when at least one constraint moment degenerates, meaning that at least one line among *gh*, *gi* and *hi* degenerates to a point. For the 3-UPU manipulator, vectors $\mathbf{n}_i$ ($i$=1,2,3) are normal to both the revolute joints axes of the U joint, i.e., $\mathbf{n}_i$ is the cross product of the unit vectors of the corresponding revolute joints axes. Accordingly, a constraint moment applied by one leg can degenerate if one of the universal joints of the leg is replaced with two revolute joints of parallel axes.

---

[1] For each leg, the axis of the U joint connected to the base platform is parallel to the one of the U joint attached to the moving platform.
[2] A joint screw stands for a twist screw associated with the joint

### B. Singularity analysis of the Delta-linear robot

The Delta-linear robot consists of three identical legs as shown in Fig. 5. The motion of the moving platform is controlled by means of three linear actuators mounted on the base and through three spatial parallelograms. Due to the arrangement of the links and joints, the manipulator performs a translational motion. In the absence of serial singularity, the transpose of the inverse Jacobian of this manipulator can be expressed as follows,

$$\mathbf{J}^{-T} = \begin{bmatrix} \hat{\mathbf{F}}_{11} & \hat{\mathbf{F}}_{12} & \hat{\mathbf{F}}_{21} & \hat{\mathbf{F}}_{22} & \hat{\mathbf{F}}_{31} & \hat{\mathbf{F}}_{32} \end{bmatrix} \tag{29}$$

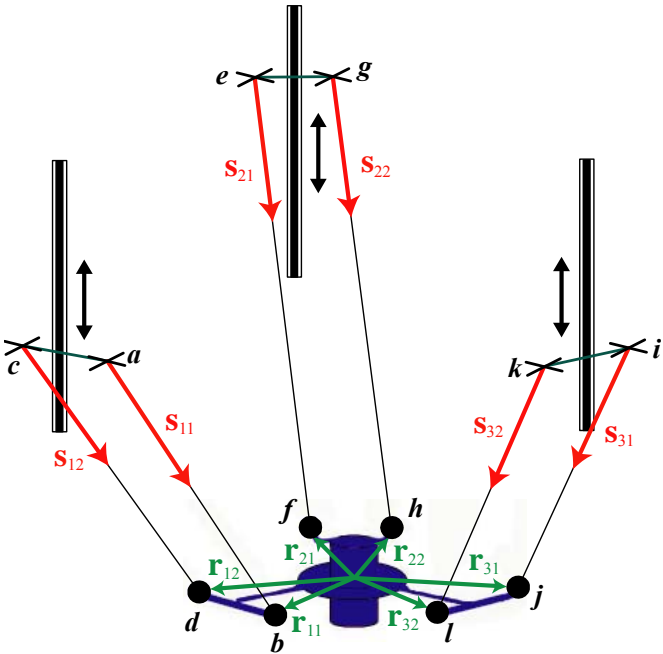

Figure 5: Delta-Linear manipulator

The columns of $\mathbf{J}^{-T}$ are zero pitch screws, $\hat{\mathbf{F}}_{ij} = \begin{bmatrix} \mathbf{s}_{ij}^T & (\mathbf{r}_{ij} \times \mathbf{s}_{ij})^T \end{bmatrix}^T$ ($i$=1,2,3, $j$=1,2), represented by lines along the six rods, $\mathbf{s}_{ij}$ being a unit vector in the direction of rod $j$ within leg $i$, and $\mathbf{r}_{ij}$ is the position vector of a point on this rod. Each screw is an actuation force, which is reciprocal to all joint screws of the $j$th rod of the $i$th leg except to the actuated prismatic joint screw of the same leg. It is noteworthy that the two actuation forces within each leg are parallel.

Furthermore, these six actuation forces form a 6-screw system, which can be replaced with another equivalent 6-screw system by applying a linear transformation to the basis of the first one [19]. Thus, each 2-system of two parallel zero pitch screws, $\hat{\mathbf{F}}_{i1} = \begin{bmatrix} \mathbf{s}_{i1}^T & (\mathbf{r}_{i1} \times \mathbf{s}_{i1})^T \end{bmatrix}^T$ and $\hat{\mathbf{F}}_{i2} = \begin{bmatrix} \mathbf{s}_{i2}^T & (\mathbf{r}_{i2} \times \mathbf{s}_{i2})^T \end{bmatrix}^T$ can be replaced by an equivalent 2-system of a zero pitch screw $\hat{\mathbf{F}}_{ei} = \hat{\mathbf{F}}_{i1}$ and infinite pitch screw $\hat{\mathbf{M}}_{ei} = \hat{\mathbf{F}}_{i1} - \hat{\mathbf{F}}_{i2} = \begin{bmatrix} \mathbf{0}_{1\times 3} & \mathbf{n}_i^T \end{bmatrix}^T$, $\mathbf{n}_i$ being the direction of the torque associated with constraint moment $\hat{\mathbf{M}}_{ei}$ and normal to the plane of the $i$th parallelogram joint.

The Delta robot, depending on the choice of its inverse Jacobian matrix, can be considered topologically equivalent to a manipulator of class I or a manipulator of class III. It is noteworthy that we obtain the same results related to singularities for both cases.

First, let us consider that this manipulator belongs to the first class of parallel manipulators. From Section IV.B, the Delta robot meets a singularity whenever ($\mathbf{s}_{11} \times \mathbf{s}_{21}$).$\mathbf{s}_{31}$=0 or ($\mathbf{n}_1 \times \mathbf{n}_2$).$\mathbf{n}_3$=0. In this case, singular configurations occur for cases (b-g) in Fig. 1. Let us notice that if case (d) (respectively, case (g)) occurs, then case (e) (respectively, case (b)) occurs necessarily. However, the reverse is not true.

Now, let us assume that the Delta-linear robot belongs to the third class of parallel manipulators. From Section IV.D, the manipulator meets a singularity whenever expression (27) is null. Since the three legs contain identical parallelograms. Therefore, **cd** = **ab** and (27) is reduced to the following expression:

$$[\mathbf{ab} \cdot \mathbf{N}][\mathbf{tu} \cdot (\mathbf{db} \times \mathbf{ab})] = 0 \tag{30}$$

with
$\mathbf{N} = \mathbf{s}_{21} \times \mathbf{s}_{31}$ and $\mathbf{tu} = (\mathbf{n}_2 \times \mathbf{n}_3)$

Accordingly, the manipulator meets a singularity whenever

$$\left[\mathbf{s}_{11} \cdot (\mathbf{s}_{21} \times \mathbf{s}_{31})\right]\left[(\mathbf{n}_2 \times \mathbf{n}_3) \cdot \mathbf{n}_1\right] = 0 \tag{31}$$

### C. Singularity analysis of the McGill SMG manipulator

The McGill Schönflies-Motion Generator (SMG) was introduced in [1,25] and is illustrated in Fig. 6. The McGill SMG is composed of two identical four-degree-of-freedom serial chains in a parallel array, sharing one common base and one common moving platform. Each chain is composed of a revolute joint, two coplanar parallelogram joints and another revolute joint. For both chains, the first two joints are actuated whereas the other two joints are passive. The whole mechanism gives, as a result, a four-degree-of-freedom motion to its end-platform, namely, three independent translations and one rotation about an axis of fixed direction.

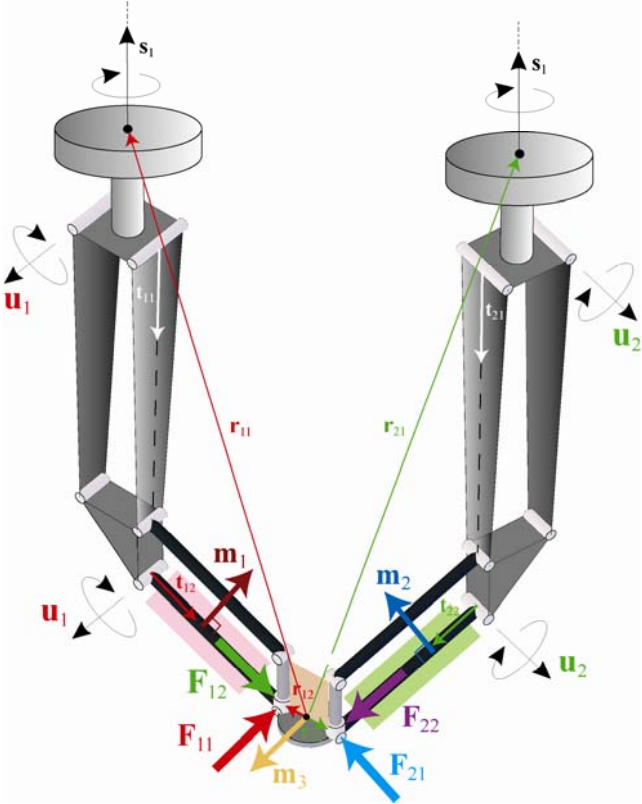

Figure 6: McGill Schönflies-Motion Generator CAD model

Thus, each leg having two actuators and a connectivity equal to four, it applies two actuation forces and two constraint moments. As the governing lines of this manipulator had never been introduced, they are presented in detail in this section.

For leg $i$, the twist screws of the two revolute joints are defined by means of the direction of their axis and the position of a point on it. Consequently, they are expressed as follows, $\hat{\mathbf{T}}_{i1} = \begin{bmatrix} \mathbf{s}_1^T & (\mathbf{r}_{i1} \times \mathbf{s}_1)^T \end{bmatrix}^T$ and $\hat{\mathbf{T}}_{i4} = \begin{bmatrix} \mathbf{s}_1^T & (\mathbf{r}_{i2} \times \mathbf{s}_1)^T \end{bmatrix}^T$, $\mathbf{s}_1$ is the rotation axis of the revolute joints and $\mathbf{r}_{i1}$ and $\mathbf{r}_{i2}$ are shown in Fig. 6. Moreover, the two coplanar parallelogram joints can only perform translation motion and their twist screws are defined as follows, $\hat{\mathbf{T}}_{i2} = \begin{bmatrix} \mathbf{0}_{1\times 3} & (\mathbf{t}_{i1} \times \mathbf{u}_i)^T \end{bmatrix}^T$ and $\hat{\mathbf{T}}_{i3} = \begin{bmatrix} \mathbf{0}_{1\times 3} & (\mathbf{t}_{i2} \times \mathbf{u}_i)^T \end{bmatrix}^T$, $\mathbf{u}_i$ being a vector normal to the parallelogram joints of leg $i$ and representing the direction of their revolute joints axes, $\mathbf{t}_{i1}$ and $\mathbf{t}_{i2}$ are the direction vector of the parallelogram links as shown in Fig. 6.

The actuation forces and constraint moments are found by applying their reciprocity to the joints screws. The actuation forces of each leg $i$ are defined as follows: $\hat{\mathbf{F}}_{i1} = \begin{bmatrix} \mathbf{u}_i^T & (\mathbf{r}_{i2} \times \mathbf{u}_i)^T \end{bmatrix}^T$ ($i$=1,2). $\mathbf{F}_{i1}$ is reciprocal to all joint screws of leg $i$ except to the actuated revolute screw, i.e., $\hat{\mathbf{T}}_{i1}$.

Likewise, $\hat{\mathbf{F}}_{i2} = \begin{bmatrix} \mathbf{t}_{i2}^T & (\mathbf{r}_{i2} \times \mathbf{t}_{i2})^T \end{bmatrix}^T$ ($i$=1,2) is reciprocal to all joint screws of leg $i$ except to the actuated parallelogram screw, i.e., $\hat{\mathbf{T}}_{i2}$. Moreover, each leg applies two constraint moments to the moving platform. These moments are characterized by the direction of their torque. Let us notice that if the latter is orthogonal to $\mathbf{s}_1$, its corresponding constraint moment is reciprocal to all joint screws. Here, we come up with four torques in the same plane, $\mathbf{s}_1$ being its normal. However, the whole mechanism is overconstrained and only two torques are necessary to represent its constraint moments. It means that we only need to choose arbitrarily two vectors orthogonal vectors to $\mathbf{s}_1$ in order to represent all the constraints applied by the legs to the platform. $\mathbf{u}_1$ and $\mathbf{u}_2$ can characterize these torques, i.e., $\hat{\mathbf{M}}_i = \begin{bmatrix} \mathbf{0}_{1\times 3} & \mathbf{u}_i^T \end{bmatrix}^T$ ($i$=1,2).

Consequently; in the absence of serial singularity, the transpose of the inverse Jacobian of this manipulator can be expressed as follows,

$$\mathbf{J}^{-T} = \begin{bmatrix} \hat{\mathbf{F}}_{11} & \hat{\mathbf{F}}_{12} & \hat{\mathbf{F}}_{21} & \hat{\mathbf{F}}_{22} & \hat{\mathbf{M}}_1 & \hat{\mathbf{M}}_2 \end{bmatrix} \quad (32)$$

From Section IV.C, $ab$, $cb$, $de$, $fe$ are the lines representing the four forces $\hat{\mathbf{F}}_{ij}$ ($i$=1,2, $j$=1,2) and $\underline{gh}$, $\underline{ih}$, are lines at infinity representing the two constraint moments $\mathbf{M}_i$ ($i$=1,2). As depicted in Fig 6, $\mathbf{m}_1$ is normal to plane ($abc$), that contains $\mathbf{F}_{11}$ and $\mathbf{F}_{12}$ and $\mathbf{m}_2$ is normal to plane ($dfe$), that contains $\mathbf{F}_{21}$ and $\mathbf{F}_{22}$. Besides, $\mathbf{m}_3$ is normal to the plane containing $\mathbf{s}_1$ and $\mathbf{r}_{12}$. From (18), the McGill SMG meets a parallel singularity whenever $(\mathbf{m}_1 \times \mathbf{m}_2) \cdot \mathbf{m}_3=0$. For instance, two parallel singular configurations are shown in Fig. 7. As a matter of fact, $\mathbf{m}_1$ and $\mathbf{m}_2$ are parallel for both singularities, i.e., their cross product is null.

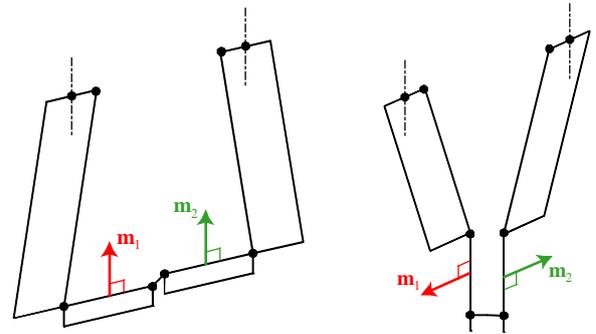

Figure 7: Two parallel singular configurations

### D. Singularity analysis of the Verne parallel module

The parallel module of the Verne machine consists of three legs, i.e., legs I, II and III. Each leg uses pairs of rods linking a prismatic joint to the moving platform through two pairs of spherical joints. Legs II and III are identical parallelograms. However, leg I differs from the other two legs in that $A_{11}A_{12} \neq B_{11}B_{12}$. Leg I does not remain planar (rod directions define skew lines) as the machine moves, unlike the other two legs that are articulated parallelograms. The motion of the moving platform is generated by the sliding of three actuators along three vertical guides [26,27]. Due to the arrangements of links and joints, the manipulator performs a complex motion defined as a simultaneous combination of translation and a slight coupled rotation as shown in Fig. 8.

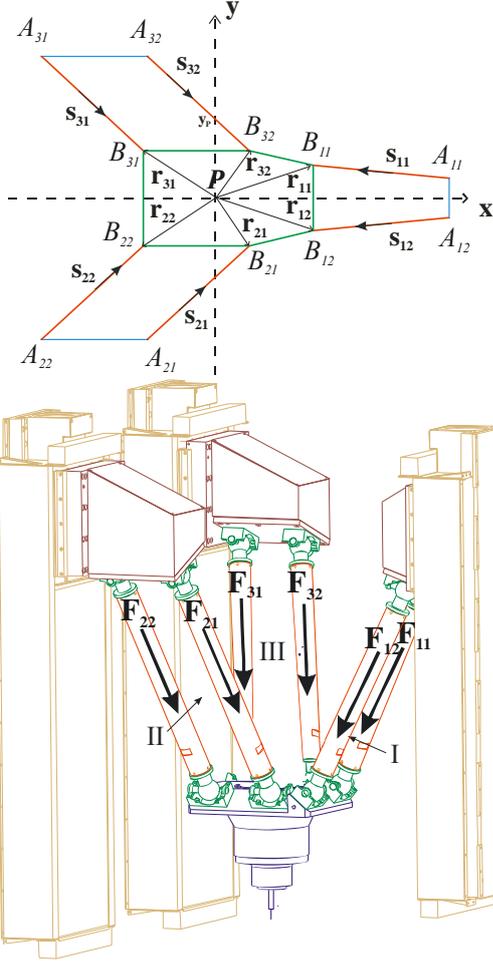

Figure 8: Schematic top view (top) and CAD representation (bottom) of the parallel Module of the Verne machine

Let us suppose that the manipulator is not in serial singularity configurations and we are only interested in studying the parallel singularities. Thus, we can consider that the transpose of the inverse Jacobian of this manipulator can be expressed as follows,

$$\mathbf{J}^{-T} = \begin{bmatrix} \hat{\mathbf{F}}_{11} & \hat{\mathbf{F}}_{12} & \hat{\mathbf{F}}_{21} & \hat{\mathbf{F}}_{22} & \hat{\mathbf{F}}_{31} & \hat{\mathbf{F}}_{32} \end{bmatrix} \quad (33)$$

The columns of $\mathbf{J}^{-T}$ are zero pitch screws, $\hat{\mathbf{F}}_{ij} = \begin{bmatrix} \mathbf{s}_{ij}^T & (\mathbf{r}_{ij} \times \mathbf{s}_{ij})^T \end{bmatrix}^T$ ($i=1,2,3$, $j=1,2$), represented by lines along the six rods, $\mathbf{s}_{ij}$ being a unit vector in the direction of rod $j$ within leg $i$ and $\mathbf{r}_{ij}$ is the position vector of a point on this rod. Each of these screws is an actuation force, which is reciprocal to all the joint screws of rod $j$ within leg $i$ except to the actuated prismatic joint screw of the same leg. It is noteworthy that the pairs of actuation forces within legs II and III are parallel, so $\mathbf{s}_{21} = \mathbf{s}_{22}$ and $\mathbf{s}_{31} = \mathbf{s}_{32}$. The parallel module of the Verne machine, having such inverse Jacobian matrix belongs to the third class of parallel manipulators, which is analyzed in Section IV.D. Here, *ab*, *cd*, *ef*, *gh*, *ij*, *kl* are the lines representing the six forces $\hat{\mathbf{F}}_{ij}$ ($i=1,2,3$, $j=1,2$) and the singular configurations occur for the six cases (*i-vi*) explained in Section IV.D and illustrated in Fig. 3b.

## VI. CONCLUSIONS

This paper introduced a methodology based on the Grassmann-Cayley Algebra to determine geometrically the singular configurations of manipulators, of which legs apply both actuation forces and constraint moments to their moving platform. Lower-mobility parallel manipulators and parallel manipulators, of which some legs do not have any spherical joint, are such manipulators. In previous works, the singularity analysis with CGA was limited to manipulators, of which legs transmit pure forces only. Moreover, the concept of infinite elements in the projective space makes it possible to analyze manipulators with pairs parallel governing lines, a feature arising in parallelogram-shaped legs. The GCA-based singularity analysis proposed in this paper should be useful at the conceptual design stage of parallel manipulators where all geometric parameters are not fully determined. In the scope of this paper, three classes of lower-mobility parallel manipulators were highlighted from the features of their governing lines. The first class includes manipulators with three actuation forces and three constraint moments. The second class includes manipulators with two pairs of intersecting actuation forces and two constraint moments. The third class includes manipulators with six actuation forces such that at least two pairs of governing lines are parallel. A physical meaning and a geometrical interpretation of singular configurations were determined for these three classes of lower-mobility parallel manipulators. finally, the singularities of four manipulators: *i*) the 3-UPU manipulator, *ii*) the Delta-linear robot, *iii*) the McGill Schönflies Motion Generator (McGill SMG) and *iv*) the parallel module of the Verne machine, were analyzed as illustrative examples. For the first two manipulators, we obtained the results presented in [8] and [28]. The singularity condition found for the McGill-SMG is apparently more general than in [1]. The singularities of the Verne machine were found for the first time in this paper.

## APPENDIX A

The 24 monomials of (13) can be expressed as follows:

$y_1 = [abcd][efgi][hjkl] \quad y_2 = -[abcd][efhi][gjkl]$

$y_3 = -[abcd][efgi][hikl] \quad y_4 = -[abcd][efhj][gikl]$

$y_5 = -[abce][dfgh][ijkl] \quad y_6 = -[abde][cfgh][ijkl]$

$y_7 = [abcf][degh][ijkl] \quad y_8 = -[abdf][cegh][ijkl]$

$y_9 = -[abce][dghi][fjkl] \quad y_{10} = [abde][cghi][fjkl]$

$y_{11} = [abcf][dghi][ejkl] \quad y_{12} = [abce][dghj][fikl]$

$y_{13} = -[abdf][cghi][ejkl] \quad y_{14} = -[abde][cghj][fikl]$

$y_{15} = [abdf][cghj][eikl] \quad y_{16} = -[abcf][dghj][eikl]$

$$y_{17} = [abcg][defi][hjkl] \quad y_{18} = -[abdg][cefi][hjkl]$$
$$y_{19} = -[abch][defi][gjkl] \quad y_{20} = -[abcg][defj][hikl]$$
$$y_{21} = [abdh][cefi][gjkl] \quad y_{22} = [abdg][cefj][hikl]$$
$$y_{23} = [abch][defj][gikl] \quad y_{24} = -[abdh][cefj][gikl]$$